\documentclass[12pt]{article}

\def\filename{\jobname.tex}

\let\simpleinput=\input

\def\input#1{%
\expandafter\let\csname prev#1\endcsname=\filename
\edef\filename{#1}
\simpleinput #1\relax
\edef\filename{
\expandafter\csname prev\filename\endcsname}}

\let\simplebibliography=\bibliography

\expandafter\ifx\csname bibliography\endcsname\relax
\else
\def\bibliography#1{\def\filename{\jobname.bbl}\simplebibliography{#1}}
\fi

\expandafter\ifx\csname AtEndDocument\endcsname\relax
\else
\AtEndDocument{\let\input=\simpleinput}
\fi

\def\space{ }
\def\pdfposition{[ @thispage /XYZ @xpos @ypos ]}

\let\preveverypar=\everypar
\newtoks\everypar
\preveverypar={
\the\everypar
}

\let\prevlabel=\label
\def\label#1{%
\prevlabel{#1}%
}

\makeatletter

\long\def\state#1#2#3#4#5{%
\def\statelabel{\label{#1}}
\def\st@atehere{#3#4}
\ifx\st@atehere\empty\expandafter\def\csname st@atementempty#1\endcsname{Y}\else#3\fi#4%
\expandafter\ifx\csname stateonce\endcsname\relax\relax\else #5\fi
\ifx#2\relax\else
\expandafter\def\csname st@atementlabel#1\endcsname{#2}
\fi
\expandafter\def\csname st@atement#1\endcsname{#4#5}
}

\newcount\restatec@unter

\def\restate#1{
\expandafter\ifx\csname stateonce\endcsname\relax
\expandafter\let\expandafter\what\csname st@atementlabel#1\endcsname
\expandafter\ifx\what\relax
\csname st@atement#1\endcsname
\else
\def\rest@teempty{Y}
\expandafter\ifx\csname st@atementempty#1\endcsname\rest@teempty
\def\statelabel{\label{#1}}
\csname st@atement#1\endcsname
\else
\expandafter\let\expandafter\counter\csname c@\what\endcsname
\restatec@unter=\counter
\counter=\expandafter\ifx\csname r@#1\endcsname\relax0\else\@kernel@ref{#1}\fi
\advance\counter by -1
\let\statelabel=\relax
\csname st@atement#1\endcsname
\counter=\restatec@unter
\fi
\fi
\fi
}

\makeatother

\def\skipif#1#2{\def\nope{\iffalse}\def\go{\iftrue}%
\if#2y\if#1\else[#1 commented out]\fi%
\expandafter\nope\else\expandafter\go\fi}
\let\skipfi=\fi

\def\draft{\iftrue\begingroup\par\em=============[draft]=============\par}
\def\enddraft{\par=============[/draft]=============\par\endgroup\fi}
\def\draft{\iffalse}
\let\enddraft=\fi

\long\def\nop#1{}

\def\l{\langle}
\def\r{\rangle}
\def\mod{M\!od}
\def\true{{\sf true}}
\def\false{{\sf false}}
\def\proof{\noindent {\sl Proof.\ \ }}
\def\qed{\hfill{\boxit{}}
  \ifdim\lastskip<\medskipamount \removelastskip\penalty55\medskip\fi}
\long\def\boxit#1{\vbox{\hrule\hbox{\vrule\kern3pt
                  \vbox{\kern3pt#1\kern3pt}\kern3pt\vrule}\hrule}}
\def\color[#1]#2{}
\def\possnewtheorem#1#2{
\expandafter\ifx\csname #1\endcsname\relax
\newtheorem{#1}{#2}
\fi
}

\possnewtheorem{theorem}{Theorem}
\possnewtheorem{corollary}{Corollary}
\possnewtheorem{lemma}{Lemma}
\possnewtheorem{definition}{Definition}
\possnewtheorem{algorithm}{Algorithm}
\possnewtheorem{counterexample}{Counterexample}

\title{Representing states in iterated belief revision}
\author{Paolo Liberatore%
\thanks{
DIAG, Sapienza University of Rome,
{\tt liberato@diag.uniroma1.it}
}
}

\begin{document}

\maketitle

\draft

FARE: separator, newpage, draft, FARE

FARE: total -> connected

FARE: anche per gli OCF esiste una rappresentazione sintattica, un insieme di
(formula,numero) invece di (mondo,numero); il punto non e' tanto la esistenza
di una rappresentazione sintattica, ma quanto la sua compattezza, che poi e'
quello che indica se e' realistica

questo articolo riassume i meccanismi: paper-5.pdf
InfOCF-Lib: A Java Library for OCF-based Conditional Inference,
Steven Kutsch

"Several semantics have been proposed for defining nonmonotonic inference
relations over sets of such rules. Examples are Lewis' system of spheres [17],
conditional objects evaluated using Boolean intervals [9], or possibility
measures [8, 10]."

"Here, we will consider Spohn's ranking functions [20] that assign degrees of
disbelief to propositional interpretations. C-Representations [13, 14] are a
subset of all ranking functions for a knowledge base, which can conveniently be
calculated by solving a constraint satisfaction problem dependent on the
knowledge base. [...] the tool InfOCF [...] allows the user to load knowledge
bases, calculate admissible ranking functions (in particular c-representations
and system Z [19]), and perform inference using these sets of ranking
functions. System P entailment [1] was also implemented."

"several new theoretical approaches have been proposed such as different
notions of minimal c-representations [2] and inference usinsets of ranking
functions [5, 7]."

FARE: altri sistemi pratici, rivedere cosa fanno di preciso:

1812.08313.pdf
Iterated belief revision under resource constraints: logic as geometry
Dan P. Guralnik, Daniel E. Koditschek

aaai.v33i01.33013076.pdf
Iterated Belief Base Revision: A Dynamic Epistemic Logic Approach
Marlo Souza, Alvaro Moreira, Renata Vieira
sembra includere lexicographic o essere equivalente

sistemi che estendono lexicographic:

- priority graph
  aaai.v33i01.33013076.pdf
  Iterated Belief Base Revision: A Dynamic Epistemic Logic Approach
  Marlo Souza, Alvaro Moreira, Renata Vieira
  2019

- combinazioni lessicografiche di ordinamenti generici
  document (11).pdf
  Operators and Laws for Combining Preference Relations
  Hajnal Andreka, Mark Ryan, Pierre-Yves Schobbens
  2002

FARE: vedere altro in iterated-revision.txt

\enddraft

\begin{abstract}

Iterated belief revision requires information about the current beliefs. This
information is represented by mathematical structures called doxastic states.
Most literature concentrates on how to revise a doxastic state and neglects
that it may exponentially grow. This problem is studied for the most common
ways of storing a doxastic state. All four methods are able to store every
doxastic state, but some do it in less space than others. In particular, the
explicit representation (an enumeration of the current beliefs) is the more
wasteful on space. The level representation (a sequence of propositional
formulae) and the natural representation (a history of natural revisions) are
more compact than it. The lexicographic representation (a history of
lexicographic revision) is even more compact than them.

\end{abstract}
\section{Introduction}

\draft

synopsis:

\begin{itemize}

\item revising requires a comparison of the strength of beliefs between the
possible situations; this is called the doxastic state;

\item the simplest and most used form of doxastic state is a connected preorder
between propositional models;

\item much of the literature is about how to change the doxastic states, but a
few studies notice that it may exponentially grow in size;

\item the most common practical representations of a doxastic state: explicit,
by levels and by history;

\item they are not the same; by increasing order of compactness: explicit,
levels and natural histories, lexicographic histories;

\item they are all universal: they express all possible connected preorders;

\item detailed description of the four representations;

\item summary of article, by sections:

\begin{itemize}

\item definitions.tex: the definition of the four representations and their
equivalence;

\item classes.tex: the equivalence classes of the connected preorder generated
by natural and lexicographic histories; this proves that that they can both
translated to a level representation;

\item comparison.tex: proves that the four representations are universal and
complete the compactness comparison;

\item related.tex: related work;

\item conclusions.tex: summary of work and future directions.

\end{itemize}

\end{itemize}

\enddraft

A train runs between Rome and Viterbo. A bus runs between Rome and Viterbo. No,
not exactly. The bus only runs when the train does not work. Something believed
may turn wrong at later time. How to revise it is belief revision.

The first studies~\cite{alch-gard-maki-85,gard-88} deemed the factual beliefs
like the train running and the bus running sufficient when facing new
information like ``either the train or the bus do not run{}''. They are
minimally changed to satisfy it. When more than one minimal change exists, they
are all equally likely.

No actual agent revises like this. The train usually runs, except when it
snows. Or not: the train line is in testing, and only runs occasionally.
Beliefs are not sufficient. Their strength is necessary.

The strength of their combinations is necessary. That the train runs while the
bus does not is more credible than the other way around, and both scenarios are
more likely that the two services being both shut down. Beliefs are not
independent to each other. Their combined strength may not be just the sum of
their individual strength. The bus running is unlikely, but almost certain when
the train does not work. The collection of this kind of information is called
doxastic state. It tells how much each possible scenario is believed to be the
case.

The simplest and most used form of doxastic state is a connected preorder
between propositional models. Each model stands for a possible situation; the
preorder says which is more believed than which.

Even in the simplest setting, propositional logic, the models are exponentially
many. Moreover, the doxastic state is not static: the revisions change it.
Scenarios conflicting with new information decrease in credibility. Scenarios
supported by new information increase. The problem of iterated belief revision
is not only how to revise a doxastic state, but also how to store it in
reasonable space. A list of comparisons ``this scenario is more credible that
this other one{}'' is always exponentially longer than the number of individual
beliefs. Exponential means intractable.

No actual agent holds an exponential amount of information. Artificial or
otherwise: computers have limited memory; people do not memorize long lists
easily.

How can a computer store a doxastic state? How do people remember a doxastic
state? Not in the form of a list. Somehow else. In a short way. Maybe by some
general rules, with exceptions.

How to store a doxastic state is not a new question. It showed up early in the
history of iterated belief revision~%
\cite{ryan-91,will-92,dixo-93,dixo-wobc-93,will-95},
resurfacing rarely from time to time
\cite{benf-etal-99,benf-etal-00,jin-thie-05,zhua-etal-07,rott-09}
until attracting interest recently
\cite{gura-kodi-18,souz-etal-19,saue-hald-19,%
schw-etal-20,arav-20,saue-beie-22,schw-etal-22}.
In spite of the many studies that concentrate on how to change the doxastic
state neglecting its size
\cite{hald-etal-21,souz-etal-21,saue-etal-22,kern-etal-22,kern-etal-23}, some
solutions already exist.

Beside listing the strength of beliefs in each possible scenario one by one,
the most common form of doxastic state is a list of logical formulae. The first
is true in the most likely scenario. The second is true in the most likely
among the remaining ones, and so on~%
\cite{will-92,dixo-93,will-95,dixo-wobc-93,%
jin-thie-05,meye-etal-02,rott-09,schw-etal-20}.

A common alternative is to not store the doxastic state itself but what creates
it. The current beliefs come from what learned in the past. The past revisions
make the doxastic state. Rather than strenuously compiling and storing the
strength of belief in every single possible scenario, it is only computed when
necessary from the sequence of previous revisions. The history is the doxastic
state~%
\cite{ryan-91,benf-etal-99,benf-etal-00,koni-pere-00,roor-etal-03,zhan-04,%
hunt-delg-07,schw-etal-22}.

Of the many ways of changing a doxastic state~\cite{rott-09}, the two most
studied ones are considered: lexicographic and natural
revision~\cite{spoh-88,naya-94,bout-96-a,chan-boot-23}. They complete the list
of the four representations compared:

\begin{description}

\item[Explicit representation:] the mathematical representation of a connected
preorder: a set of pairs of models; each model describes a possible scenario, a
pair expresses a stronger belief in the first than in the second; every
doxastic state can be represented by such a set or pairs, which may however be
very large;

\item[Level representation:] a sequence of formulae; the first describes the
most strongly believed scenarios, the second describes the most strongly
believed remaining ones, and so on; these sequences represent every doxastic
state, and may do that in less space than in the explicit representation;

\item[Histories of natural revisions:] they represent all doxastic states; they
can be converted into the level representation and back without an exponential
growth;

\item[Histories of lexicographic revisions:] they represent all doxastic
states, and do that in the most space-saving way among the four considered
methods: the others can be converted into lexicographic histories with a
limited increase of size, while the inverse translation may not.

\end{description}

\noindent
\begin{figure}
\begin{center}
\setlength{\unitlength}{3750sp}%
\begin{picture}(1980,2044)(4411,-5834)
\thinlines
{\color[rgb]{0,0,0}\put(5251,-5611){\vector(-1, 1){675}}
}%
{\color[rgb]{0,0,0}\put(5551,-5611){\vector( 1, 1){675}}
}%
{\color[rgb]{0,0,0}\put(4538,-4673){\vector( 1, 1){675}}
}%
{\color[rgb]{0,0,0}\put(6188,-4673){\vector(-1, 1){675}}
}%
{\color[rgb]{0,0,0}\put(4726,-4861){\vector( 1, 0){1350}}
}%
{\color[rgb]{0,0,0}\put(6076,-4711){\vector(-1, 0){1350}}
}%
\put(5401,-5761){\makebox(0,0)[b]{\smash{\fontsize{9}{10.8}
\usefont{T1}{cmr}{m}{n}{\color[rgb]{0,0,0}$explicit$}%
}}}
\put(4426,-4861){\makebox(0,0)[b]{\smash{\fontsize{9}{10.8}
\usefont{T1}{cmr}{m}{n}{\color[rgb]{0,0,0}$level$}%
}}}
\put(6376,-4861){\makebox(0,0)[b]{\smash{\fontsize{9}{10.8}
\usefont{T1}{cmr}{m}{n}{\color[rgb]{0,0,0}$natural$}%
}}}
\put(5401,-3961){\makebox(0,0)[b]{\smash{\fontsize{9}{10.8}
\usefont{T1}{cmr}{m}{n}{\color[rgb]{0,0,0}$lexicographic$}%
}}}
\end{picture}%
\nop{
         lexicographic
          ^         ^
          |         |
          |         |
          | ------> |
      levels       natural
          ^ <------ ^
          |         |
          |         |
          |         |
           explicit
}
\caption{Comparison of the four considered representations}
\label{figure-comparison}
\end{center}
\end{figure}
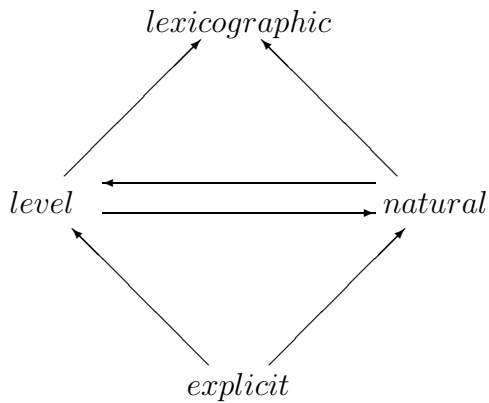

Figure~\ref{figure-comparison} shows the existence of polynomially-bounded
translations between the four considered representations.

The following is a summary of the article section by section.

Section~\ref{definitions} formally defines the four considered representations
of doxastic states and formalizes their equivalence.

Section~\ref{classes} shows the equivalence classes of the connected preorders
resulting from natural and lexicographic revisions. This proves that the
inductive definitions of the previous section match the usual definitions in
terms of equivalence classes, and also proves that both representations are
translatable into the level representation.

Section~\ref{comparison} shows that the considered representations are
universal, and completes their space efficiency comparison. Lexicographic
histories are strictly more compact than the level and natural histories
representations, which compare the same and are strictly better than the
explicit representation.

A comparison with the related literature in Section~\ref{related}, followed by
a discussion of the results and the possible future directions of study in
Section~\ref{conclusions}.

\draft

histories of other operators such as restrained revision~%
\cite{boot-meye-06,rott-09};
other subcases of propositional logics~\cite{krom-67,post-41};
equivalence and redundancy of natural revision histories, or histories of mixed
revisions~\cite{libe-23};
measures other than the size of doxastic states such as the complexity of
performing a further revision~\cite{libe-97,schw-etal-20};
doxastic states in other representations such as sets of
conditionals~\cite{gura-kodi-18}
or more complex than an ordering between models such as ordinal conditional
functions~\cite{kuts-19}, proper ordinal intervals~\cite{boot-chan-20},
priority graphs~\cite{liu-11,souz-etal-19} or combinations of other
orderings~\cite{andr-etal-02}.

\enddraft

\section{Definitions}
\label{definitions}

Doxastic states may take several forms, such as connected preorders, rankings
and systems of spheres~\cite{ferm-hans-11}. Connected preorders are studied in
this article: an order between propositional models that is reflexive ($I \leq
I$), transitive ($I \leq J$ and $J \leq H$ entail $I \leq H$) and connected
(either $I \leq J$ or $J \leq I$).

Epistemically, $I \leq J$ means that $I$ is a more believed scenario than $J$.

Mathematically, an order between propositional models is a set of pairs of
models: it contains $\l I,J \r$ if $I$ is less than or equal than $J$, or $I
\leq J$. This explicit representation of an order may take space linear in the
number of models, which is exponential in the number of variables in
propositional logics.

The alternative representations of an ordering considered in this article are
by a formula for equivalence class and by a sequence of lexicographic or
natural revisions. They all comprise a sequence of formulae. This calls for a
wording simplification, where a sequence is identified with the order it
represents. For example, ``the lexicographic order $[S_1,\ldots,S_m]$'' is the
order that results from the sequence of lexicographic revisions
$S_m,\ldots,S_1$ starting from a void doxastic state, where models are all
equally believed.

The same sequence has different meanings in different representations: the
natural order $[a \vee b,\neg a]$ differs from the lexicographic order $[a \vee
b,\neg a]$. In the other way around, the same order is given by different
sequences in different representation. For example, the lexicographic order
$[a,b]$ is the same as the level order
{} $[a \wedge b, a \wedge \neg b, \neg a \wedge b, \neg a \wedge \neg b]$.
Technically, they are equivalent: $I \leq J$ holds in the first if and if it
holds in the second. Equivalence is equality of $I \leq J$ for all pair of
models.

\subsection{The explicit order}

The explicit order is the mathematical definition of an order between
propositional models: a set of pairs of models. The set of all propositional
models over the given alphabet is denoted by $M$.

\begin{definition}
\label{explicit}

The {\em explicit order} induced by $S \subseteq M \times M$ compares $I \leq_S
J$ if $\l I,J \r \in S$, where $M$ is the set of all models.

\end{definition}

A connected preorder is reflexive, transitive and connected:

\begin{description}

\item[reflexive:] $\l I,I \r \in S$ for every $I \in M$;

\item[transitive:] $\l I,J \r \in S$ and $\l J,H \r \in S$ imply $\l I,H \r \in
S$ for every $I,J,H \in M$;

\item[connected:] either $\l I,J \r \in S$ or $\l J,I \r \in S$ for every $I,J
\in M$.

\end{description}

A connected preorder $S \subseteq M \times M$ is the same as a sequence of
disjoint sets of models $S=[S_1,\ldots,S_m]$, where every element $S_i$ is a
set of models: $S_i \subseteq M$ and $S_i \cap S_j$ if $i \not= j$. The
correspondence is:

\begin{itemize}

\item $I \leq_S J$ if and only if $I \in M_i$, $J \in M_j$ and $i \leq j$;

\item $S_i = \{I \in M \backslash (S_1 \cup \cdots \cup S_{i-1})
\mid \forall J \in M \backslash (S_1 \cup \cdots \cup S_{i-1}) ~.~ I \leq J\}$

\end{itemize}

The first set $S_1$ comprises all minimal models according to $\leq$. The
second set $S_2$ comprises the minimal models except for $S_1$. They are
minimal among what remains: $M \backslash S_1$.

\draft

\[
\l I,J \r \in S
\]

\enddraft

\subsection{The level order}

The explicit order takes quadratic space in the number of models, which is
exponential in the number of variables. Space can often be significantly
reduced by turning every set $M_i$ into a propositional formula. This is the
most used realistic representation of a doxastic state in iterated belief
revision~%
\cite{will-92,dixo-93,will-95,dixo-wobc-93,%
jin-thie-05,meye-etal-02,rott-09,schw-etal-20}.

\begin{definition}[Level order]
\label{level}

The {\em level order} induced by the sequence of formulae $S =
[S_1,\ldots,S_m]$ that are mutually inconsistent and whose disjunction is
tautological compares $I \leq_S J$ if $i \leq j$ with $I \models S_i$ and $J
\models S_j$.

\end{definition}

Variants lift the condition of mutual inconsistency or tautological disjunction
or add the requirement of no single inconsistent formula. In the first, $i$ and
$j$ are the minimal indexes of formulae satisfied by $I$ and $J$. In the second,
the definition is added ``or $J$ does not satisfy any formula of $S$''. The
third does not require any modification. These changes are inessential:

\begin{enumerate}

\item the order $[S_1,\ldots,S_m]$ is the same as
{} $[S_1,
{}   S_2 \wedge \neg S_1,
{}   \ldots,
{}   S_m \wedge \neg S_{m-1} \wedge \cdots \wedge S_1]$,
which comprises mutually inconsistent formulae;

\item the order $[S_1,\ldots,S_m]$ is the same as
$[S_1,\ldots,S_m,\neg S_1 \vee \cdots \vee \neg S_m]$,
whose disjunction of formulae is tautological;

\item the order $[S_1,\ldots,S_m]$ is the same as the same sequence with all
inconsistent formulae removed.

\end{enumerate}

\draft

\[
i \leq j \mbox{ where } I \models S_i \mbox{ and } J \models S_j
\]

\enddraft

\subsection{The lexicographic order}

The lexicographic order is what results from a sequence of lexicographic
revisions~\cite{spoh-88,naya-94} applied to a void doxastic state, where all
models compare the same. A number of other iterated revision operators have
been proved to be reducible to it~\cite{libe-23}, making it a good candidate
for representing arbitrary doxastic states.

The first step in the definition of this order is the order induced by a single
formula: believing a formula is the same as believing that every scenario where
it is true is more likely than every scenario where it is false.
Mathematically, its models are less than the others.

\begin{definition}
\label{order-formula}

The {\em order induced by a formula} $F$ compares $I \leq_F J$ if either $I
\models F$ or $J \not\models F$.

\end{definition}

The definition implies that $I \leq_F J$ holds in exactly three cases:

\begin{itemize}

\item $I \models F$ and $J \not\models F$ (strict order),

\item $I \models F$ and $J \models F$ (equivalence, first case), or

\item $I \not\models F$ and $J \not\models F$ (equivalence, second case).

\end{itemize}

The principle of the lexicographic order is that the last-coming formula makes
the bulk of the ordering, separating its satisfying models from the others. The
previous formulae matter only for ties. The following definition applies this
principle to the condition $I \leq_S J$, where $S=[S_1,\ldots,S_m]$ is a
sequence of lexicographic revisions in reverse order: $S_m$ is the first, $S_1$
the last.

\begin{definition}
\label{lexicographic}

The {\em lexicographic order} induced by the sequence of formulae $S =
[S_1,\ldots,S_m]$ compares $I \leq_S J$ if

\begin{itemize}

\item either $S=[]$ or

\item
\begin{itemize}

\item[] $I \leq_{S_1} J$ and

\item[]
either $J \not\leq_{S_1} I$
	or
$I \leq_R J$, where $R = [S_2,\ldots,S_m]$.

\end{itemize}

\end{itemize}

\end{definition}

The sequence $S$ is identified with the order, giving the simplified wording
``the lexicographic order $S$''.

The lexicographic order $I \leq_S J$ is equivalently defined in terms of the
strict part and the equivalence relation of $\leq_F$:

\begin{itemize}

\item either $I <_{S_1} J$, or

\item
$I \equiv_{S_1} J$
	and 
$I \leq_R J$, where $R = [S_2,\ldots,S_m]$.

\end{itemize}

\draft

\[
I \leq_F J \mbox{ is }
I \models F \mbox{ or } J \not\models F
\]

\[
S=[]
\mbox{ or }
(
	I \leq_{S_1} J
	\mbox{ and }
	(
		J \not\leq_{S_1} I
		\mbox{ or }
		I \leq_R J
	)
)
\]

\enddraft

\subsection{The natural order}

Like the lexicographic order is what results from a sequence of lexicographic
revisions, every other revision gives a way to represent an ordering. One early
and much studied such operator is the natural revision~\cite{spoh-88,naya-94}.
Along with lexicographic and restrained revision is one of the three elementary
revision operators~%
\cite{chan-boot-23}.

The founding principle of natural revision is to alter the doxastic state as
little as possible to make the revising formula believed. A scenario becomes
believed when it is one of the most believed scenario according to the
formulae. The comparison is otherwise unchanged.

\begin{definition}[Natural order]
\label{natural}

The natural order induced by the sequence of formulae $S = [S_1,\ldots,S_m]$
compares $I \leq_S J$ if either $S = []$ or:

\begin{itemize}

\item
$I \in \mod(S_1)$
	and
$\forall K \in \mod(S_1) . I \leq_R K$, or

\item
$I \leq_R J$
	and
	either
	$J \not\in \mod(S_1)$
		or
	$\exists K \in \mod(S_1) . J \not\leq_R K$,

\end{itemize}

\noindent where $R = [S_2,\ldots,S_m]$.

\end{definition}

The simplified wording ``the natural order $S$'' stands for the order induced
by the sequence $S$.

Being $\leq_R$ a connected preorder, the recursive subcondition $J \not\leq_R
K$ is the same as $K <_R J$.

This definition implements its justification: that the minimal models of the
revising formula are made minimal while not changing the relative order among
the others. The next section will prove it by expressing the natural order on
equivalence classes.

\draft

follows a sketch of a proof; is long, and the following section contains the
formal definition, making it mostly useless

\enddraft

\skipif{}{y}

\begin{itemize}

\item the first condition says that $I \leq_S J$ holds if $I$ has been lowered
by the last revision

this happens if $I$ satisfies $S_1$ and is less than or equal than all other
models of $S_1$

this is the definition of I<=J; it does not matter whether J has been lowered
as well; that would only imply J<=I as well, but I<=J still holds; this the two
conditions of I <=S J are disjoined, the second does not matter if the first
holds

\item the second condition says that I<=J holds if it did and J has not been
lowered by the last revision

the clause "... unless $I$ has been lowered as well" could be added, but it is
not necessary; if I had been lowered, the first condition holds, which makes
this second one irrelevant; therefore, I can be assumed not lowered by the last
revision

the order I <=S J did hold is the same as I <=R J

J is lowered when it is in $S_1$ and is minimal:

\begin{itemize}
\item $J \in S_1$ and
\item $\not\exists K \in \mod(S_1) . K <_R J$
\end{itemize}

the converse is obtained by applying demorgan:

\begin{itemize}
\item $J \not\in S_1$ or
\item $\exists K \in \mod(S_1) . K <_R J$
\end{itemize}

since the order is connected, K <R J is the same as J /<=R K

\end{itemize}

this ends the explanation/proof that the definition generates the order
obtained by the natural revision

\skipfi

\draft

\[
S = []
\mbox{ or }
(I \in \mod(S_1) \mbox{ and } \forall K \in \mod(S_1) . I \leq_R K)
\mbox{ or }
(
	I \leq_R J
	\mbox{ and }
		(
		J \not\in \mod(S_1)
		\mbox{ or }
		\exists K \in \mod(S_1) . J \not\leq_R K
		)
)
\]

\enddraft

\subsection{Different sequences, same order}

The same order can be represented in different ways. The explicit order $S$ may
be the same as the level order $R$, the lexicographic orders $Q$ and $T$ and
the natural order $V$. Same order, different representations or different
sequences in the same representation. These sequences are equivalent on their
induced order.

\begin{definition}[Equivalence]
\label{equivalence}

Two orders $S$ and $R$ are equivalent, denoted $S \equiv R$, if $I \leq_S J$
and $I \leq_R J$ coincide for all pairs of models $I$ and $J$.

\end{definition}

This definition allows writing ``the level order $R$ is equivalent to the
lexicographic order $Q$ and to the lexicographic order $T$'', meaning that the
three sequences $R$, $Q$ and $T$ represent the same order.

Such statements are used when comparing two different representations, like
when proving that every natural order $S$ is equivalent to a lexicographic
order $R$.

Sometimes, non-equivalence is easier to handle than equivalence: two sequences
$S$ and $R$ are not equivalent if $I \leq_S J$ and $I \not\leq_R J$ or the same
with $S$ and $R$ swapped for some models $I$ and $J$. The same conditions with
$I$ and $J$ swapped is not necessary because $I$ and $J$ are arbitrary.

\draft

summary of the considered problems is in the introduction

since this cut of the article only contains the problems of universality of
polynomial-size translatability between different semantics, this list is no
longer needed

\enddraft

\skipif{}{y}
(subsection)

\subsection{Next}

The formal definitions allow for formalizing the considered problems:

\begin{enumerate}

\item are the revision-based definitions equivalent to the commonly-given ones
based on equivalence classes?

\item which representations are universal~\cite{gura-kodi-18,schw-etal-22}?
does every arbitrary connected preorder have a level, lexicographic or natural
representation?

\item which representation is more compact~%
\cite{mccl-56,rude-sang-87,coud-94,theo-etal-96,coud-sasa-02,uman-etal-06,%
cado-etal-99,cado-etal-00,eite-wang-08,endr-etal-16,farg-meng-21}?
a representation is better than another if it is able to represent the same
order in less space; formally, for every order in the first representation
there is an equivalent order in the second that takes comparable (polynomial)
amount of space, but not the other way around.

\end{enumerate}

\skipfi

\section{Classes}
\label{classes}

Four representations of doxastic states are given in the previous section:
explicit, level, lexicographic and natural. They are all defined in terms of
whether $I \leq J$ holds or not.

Iterated belief revision is often defined in terms of how they change the
doxastic state expressed in terms of its equivalence classes~\cite{rott-09}.
For example, natural revision moves the models of the first class having models
consistent with the new information to a new, first equivalence class.

A sequence of equivalence classes is the same as the level order. These
definitions say how lexicographic and natural revision change a level order.
This section does that for the definitions in the previous section. It shows
how to translate lexicographic and natural orders into level orders. This also
proves that the definitions of the lexicographic and natural revisions match
the definitions in terms of equivalence classes from the literature.

The proof scheme is:

\begin{itemize}

\item the natural order $[]$ is equivalent to the level order $[\true]$ since
$I \leq J$ holds for all models in both;

\item the two orders are kept equivalent while adding formulae at the front of
the natural order; this requires:

\begin{itemize}

\item showing the order resulting from adding a single formula in front of the
natural order;

\item expressing that order in the level representation.

\end{itemize}

\end{itemize}

The lexicographic representation is treated similarly.
Details are in Appendix~\ref{proofs}.

\subsection{From natural to level orders}

The reduction from natural orders to level orders follows the scheme outlined
above: the base case is a correspondence between the level order $[\true]$ and
the natural order $[]$; the induction case maintains the correspondence while
adding a formula at time to the natural order. The first proof step shows the
correspondence for the natural order $[]$.

\state{natural-empty}{lemma}{}{

\begin{lemma}
\statelabel

The natural order $[]$ is equivalent to the level order $[\true]$.

\end{lemma}

}{

\proof The definition of the level order is satisfied by every pair of models
$I,J$ since both models satisfy the first formula of the level order $[\true]$;
as a result, both $i$ and $j$ are equal to $1$, and they therefore satisfy $i
\leq j$.

The definition of the natural order is satisfied because of its first part $S =
[]$.~\qed

}

The induction step starts from the equivalence of a natural and a level order
and maintains the equivalence while adding a formula at time at the front of
the natural order, which is the same as a single natural revision.

This requires a property of natural orders: the models of the first formula
that is consistent with a new formula are also the minimal models of the new
formula according to the order.

\state{first-consistent}{lemma}{}{

\begin{lemma}
\statelabel

If $Q_c$ is the first formula of the level order $Q$ that is consistent with
the formula $S_1$, then
{} $I \models S_1 \wedge Q_c$
is equivalent to
{} $I \in \mod(S_1) \mbox{ and } \forall K \in \mod(S_1) . I \leq_Q K$.

\end{lemma}

}{

\proof The two cases are considered in turn: either
{} $I \models S_1 \wedge Q_c$
holds or it does not.

\begin{itemize}

\item $I \models S_1 \wedge Q_c$


The claim is made of two parts:
{} $I \in \mod(S_1)$
and 
{} $\forall K \in \mod(S_1) . I \leq_Q K$.

The first part $I \in \mod(S_1)$ holds because $I$ satisfies $S_1 \wedge Q_c$.

The second part
{} $\forall K \in \mod(S_1) . I \leq_Q K$
is now proved.

Let $Q_i$ be the only formula of $Q$ such that $I \models Q_i$. The assumption
$I \models Q_c$ implies $c = i$.

Let $K$ be an arbitrary model of $S_1$. The claim is $I \leq_Q K$.

Only one formula $S_k$ is satisfied by $K$. Since $Q$ satisfies $S_1$ and
$Q_k$, it also satisfies $S_1 \wedge Q_k$. As a result, $S_1 \wedge Q_k$ is
consistent.

Since $Q_c$ is the first formula of $S$ that is consistent with $S_1$, and
$Q_k$ is also consistent with $S_1$, it holds $c \leq k$. The equality of $c$
and $i$ proves $i \leq k$, which defines $I \leq_Q K$.

\item $I \not\models S_1 \wedge Q_c$


The claim is that either
{} $I \in \mod(S_1)$
or
{} $\forall K \in \mod(S_1) . I \leq_Q K$
is false.

If $I \not\models S_1$ the first part of this condition is falsified and the
claim is therefore proved. It remains to be proved when $I \models S_1$.

Let $Q_i$ be the only formula satisfied by $I$. Since $I$ satisfies $S_1$, it
also satisfies $S_1 \wedge Q_i$. This formula is therefore satisfiable. Since
$Q_c$ is the first formula of the sequence that is consistent with $S_1$, the
index $c$ is less than or equal than the index $i$. If $c$ were equal to $i$,
then $S_1 \wedge Q_c$ would be $S_1 \wedge Q_i$. Yet, $I$ is proved to satisfy
the latter and assumed not to satisfy the former. The conclusion is $c<i$.

Since $Q_c$ is by assumption the first formula of $Q$ that is consistent with
$S_1$, the conjunction $S_1 \wedge Q_c$ is consistent. Let $K$ be one of its
models. Let $Q_k$ be the only formula of $Q$ that $K$ satisfies. Since $K$
satisfies $S_1 \wedge Q_c$, it also satisfies $Q_c$. As a result, $k$ coincides
with $c$.

The conclusions of the last two paragraphs $c<i$ and $k=c$ imply $k < i$. This
is the opposite of $i \leq k$, which defined $I \leq_Q K$. The conclusion is
that a model $K$ of $S_1$ that falsifies $I \leq_Q K$ exists. This proves the
falsity of
{} $\forall K \in \mod(S_1) . I \leq_Q K$,
as the claim requires.

\end{itemize}~\qed

}

The following lemma shows how a single natural revision by a formula $S_1$
changes a level order $Q$. Technically, "a single natural revision" is
formalized as an addition to the front of a natural order. Since a natural
order is a sequence of natural revisions, revising $[S_2,\ldots,S_m]$ makes it
$S=[S_1,S_2,\ldots,S_m]$. The lemma expresses $I \leq_S J$. The expression is
in terms of a natural order $[S_2,\ldots,S_m]$ equivalent to a level order $Q$.
In other words, it tells $I \leq_S J$ where $S$ is equivalent to naturally
revising $Q$ by a formula $S_1$.

\state{natural-next}{lemma}{}{

\begin{lemma}
\statelabel

If $S=[S_1,S_2,\ldots,S_m]$ is a natural order, $Q$ is a level order equivalent
to the natural order $[S_2,\ldots,S_m]$ and $Q_c$ is the first formula of $Q$
that is consistent with $S_1$, then $I \leq_S J$ is:

\begin{itemize}

\item true if $I \models S_1 \wedge Q_c$;

\item false if $I \not\models S_1 \wedge Q_c$ and $J \models S_1 \wedge Q_c$;

\item same as $I \leq_Q J$ otherwise.

\end{itemize}

\end{lemma}

}{

\proof The definition of natural order is that $I \leq_S J$ holds if and only
if:

\begin{eqnarray*}
&& S = [] \mbox{ or } \\
&&
(I \in \mod(S_1) \mbox{ and } \forall K \in \mod(S_1) . I \leq_R K)
\mbox{ or } \\
&&
	I \leq_R J
	\mbox{ and }
		(
		J \not\in \mod(S_1)
		\mbox{ or }
		\exists K \in \mod(S_1) . J \not\leq_R K
		)
)
\end{eqnarray*}

Since the statement of the lemma predicates about a first formula $S_1$ of $S$,
whose existence falsifies the first part of this definition. The statement also
assumes that a formula of $Q$ is consistent with $S_1$, which proves its
satisfiability. The statement also assumes that $\leq_R$ is the same as
$\leq_Q$. The definition of natural order therefore becomes:

\begin{eqnarray*}
&&
(I \in \mod(S_1) \mbox{ and } \forall K \in \mod(S_1) . I \leq_Q K)
\mbox{ or }
\\
&&
(
	I \leq_Q J
	\mbox{ and }
		(
		J \not\in \mod(S_1)
		\mbox{ or }
		\exists K \in \mod(S_1) . J \not\leq_Q K
		)
)
\end{eqnarray*}

The three cases are considered in turn.


\begin{description}

\item[$I \models S_1 \wedge Q_c$]

Lemma~\ref{first-consistent} proves that $I \models S_1 \wedge Q_c$ implies
{} $I \in \mod(S_1) \mbox{ and } \forall K \in \mod(S_1) . I \leq_Q K$.
This is the first part of the rewritten definition of $I \leq_S J$, which
therefore holds.

\item[$I \not\models S_1 \wedge Q_c$ e $J \models S_1 \wedge Q_c$]

Lemma~\ref{first-consistent} proves that $I \not\models S_1 \wedge Q_c$ implies
that
{} $I \in \mod(S_1) \mbox{ and } \forall K \in \mod(S_1) . I \leq_Q K$
is false. This is the first part of the rewritten definition of $I \leq_S J$,
which is therefore equivalent to its second part:

\[
I \leq_Q J
\mbox{ and }
(J \not\in \mod(S_1) \mbox{ or } \exists K \in \mod(S_1) . J \not\leq_Q K)
\]

Lemma~\ref{first-consistent} also applies to $J \models S_1 \wedge Q_c$ with
$J$ in place of $I$. It proves
{} $J \in \mod(S_1) \mbox{ and } \forall K \in \mod(S_1) . J \leq_Q K$.
Its negation is therefore false:
{} $J \not\in \mod(S_1) \mbox{ or } \exists K \in \mod(S_1) . J \not\leq_Q K$.
This is the second part of the rewritten definition of $I \leq_S J$, which is
therefore false, as the conclusion of the lemma requires in this case.

\item[$I \not\models S_1 \wedge Q_c$ e $J \not\models S_1 \wedge Q_c$]

As proved above, the assumption $I \not\models S_1$ transforms the definition
of $I \leq_S J$ into
{} $I \leq_Q J \mbox{ and }
{} (J \not\in \mod(S_1) \mbox{ or } \exists K \in \mod(S_1) . J \not\leq_Q K)$.

Lemma~\ref{first-consistent} applies to $J \models S_1 \wedge Q_c$ with $J$ in
place of $I$. It proves that $J \not\models S_1 \wedge Q_c$ implies the falsity
of
{} $J \in \mod(S_1) \mbox{ and } \forall K \in \mod(S_1) . J \leq_Q K$.
This condition is the second part of the rewritten definition of $I \leq_S J$,
which is therefore equivalent to the first part $I \leq_Q J$, as the conclusion
of the lemma requires in this case.

\end{description}~\qed

}

The plan of the proof is to start from the equivalent natural order $[]$ and
level order $[\true]$ and to keep adding a single formula at time to the front
of the first while keeping the second equivalent to it. This requires the level
order that results from applying a natural revision. The previous lemma shows
the order in terms of a condition equivalent to $I \leq J$. The following shows
this order in the level representation.

\state{level-natural-level}{lemma}{}{

\begin{lemma}
\statelabel

If the natural order $[S_2,\ldots,S_m]$ is equivalent to the level order
$Q=[Q_1,\ldots,Q_k]$, then the natural order $S=[S_1,S_2,\ldots,S_m]$ is
equivalent to the level order $R = [R_1,R_2,\ldots,R_{k+1}]$, where
{} $R_1 = S_1 \wedge Q_c$,
{} $R_i = \neg R_1 \wedge Q_{i-1}$ for every $i > 1$
and $Q_c$ is the first formula of $Q$ such that $S_1 \wedge Q_c$ is consistent.

\end{lemma}

}{

\proof The sequence $R$ is proved to be a level order and the disjunction of
all its formulae to be tautologic. The sequence $Q$ has the same properties by
assumption. Each formula $R_i$ with $i > 1$ is $\neg R_1 \wedge Q_i$. Its
models are the models of $Q_i$ minus some models. Since $Q_i$ and $Q_j$ do not
share models, $R_i$ and $R_j$ do not either. Since the models subtracted from
each $Q_i$ when forming $R_i$ are moved to $R_1$, which is also in $R$, the
union of the models of $R$ is exactly the union of the models of $Q$, the set
of all models.

Lemma~\ref{natural-next} proves that $I \leq_S J$ is:

\begin{itemize}

\item true if $I \models S_1 \wedge Q_c$;

\item false if $I \not\models S_1 \wedge Q_c$ and $J \models S_1 \wedge Q_c$;

\item same as $I \leq_Q J$ otherwise.

\end{itemize}

The proof shows that $\leq_R$ has the same values in the same cases.

\begin{description}

\item[$I \models S_1 \wedge Q_c$]

The first formula of $R$ is $R_1 = S_1 \wedge Q_c$ by definition. Since $I$
satisfies $S_1 \wedge Q_c$ by assumption, it satisfies $R_1$. This proves that
$I \models R_i$ with $i = 1$.

Let $R_j$ be the formula of $R$ such that $J \models R_j$. Since indexes start
at $1$, it holds $1 \leq j$. The equality $i = 1$ proves $i \leq j$. The claim
$I \leq_R J$ follows.

\item $I \not\models S_1 \wedge Q_c$ and $J \models S_1 \wedge Q_c$

As in the previous case, $J \models S_1 \wedge Q_c$ implies $J \models R_j$
with $j = 1$.

Let $R_i$ be the formula of $R$ satisfies by $I$. If $i$ were $1$, then $I$
would satisfy $R_1$, which is $S_1 \wedge Q_c$. Since $I$ does not satisfy this
formula by assumption, $i$ is not $1$. Since indexes start at one, $i$ is
strictly greater than one: $i > 1$.

The conclusions $j = 1$ and $i > 1$ prove $i > j$, which is the exact opposite
of $i \leq j$. The claimed falsity of $I \leq_R J$ is therefore proved.

\item $I \not\models S_1 \wedge Q_c$ and $J \not\models S_1 \wedge Q_c$

Let $R_i$ and $R_j$ be the formulae respectively satisfied by $I$ and $J$.
Since neither model satisfies $R_1 = S_1 \wedge Q_c$ by assumption, both $i$
and $j$ are strictly greater than one: $i > 1$ and $j > 1$.

The formulae $R_i$ and $R_j$ for indexes greater than one are respectively
defined as $\neg R_1 \wedge Q_{i-1}$ and $\neg R_1 \wedge Q_{j-1}$. Since $I$
and $J$ respectively satisfy them, they respectively satisfy $Q_{i-1}$ and
$Q_{j-1}$. The level order $I \leq_Q J$ is therefore equivalent to $i - 1 \leq
j - 1$, which is the same as $i \leq j$. This is also the definition of $I
\leq_S J$.

This proves the claimed equality of $I \leq_S J$ and $I \leq_Q J$.

\end{description}~\qed

}

The two requirements for induction are met: the base step by
Lemma~\ref{natural-empty} and the induction step by
Lemma~\ref{level-natural-level}. Starting from the natural order $[]$ and the
level order $[\true]$ and adding $S_m$, then $S_{m-1}$, and continuing until
$S_1$ to the first results in a level order equivalent to
$[S_1,\ldots,S_{m-1},S_m]$.

This is enough for translating a natural order into a level order of comparable
size. Yet, it is not the way natural revision is normally expressed in terms of
equivalence class. That would prove that the definition of natural revision of
the previous section matches that commonly given. The following theorem
provides that.

\state{levels-natural-levels-poly}{theorem}{}{

\begin{theorem}
\statelabel

If the natural order $[S_2,\ldots,S_m]$ is equivalent to the level order
$Q=[Q_1,\ldots,Q_k]$, then the natural order $S=[S_1,S_2,\ldots,S_m]$ is
equivalent to the following level order $R$, where $Q_c$ is the first formula
of $Q$ that is consistent with $S_1$.

\[
R = [S_1 \wedge Q_c,
Q_1, \ldots, Q_{c-1},
\neg S_1 \wedge Q_c,
Q_{c+1}, \ldots, Q_k]
\]

\end{theorem}

}{

\proof Lemma~\ref{level-natural-level} proves that the natural order $S$ is
equivalent to the level order $R = [R_1,R_2,\ldots,R_{k+1}]$, where
{} $R_1 = S_1 \wedge Q_c$
and
{} $R_i = \neg R_1 \wedge Q_{i-1}$ for every $i > 1$.

This is the same sequence as in the statement of the lemma.




The first formula is the same in both sequences: $S_1 \wedge Q_c$.

The formula $R_i$ for $i=c+1$ in the sequence of the lemma is also the same as
the formula in the sequence of the theorem. Since $R_1$ is $S_1 \wedge Q_c$,
the formula
{} $R_i = \neg R_1 \wedge Q_{i-1}$ for $i=c+1$
in the sequence of the lemma is the same as
{} $\neg (S_1 \wedge Q_c) \wedge Q_c$,
which is equivalent to
{} $(\neg S_1 \vee \neg Q_c) \wedge Q_c$,
in turn equivalent to
{} $(\neg S_1 \wedge Q_c) \vee (\neg Q_c \wedge Q_c)$
and to
{} $\neg S_1 \wedge Q_c$,
which is the formula in the sequence of the theorem.

The formulae $R_i$ with $i$ strictly greater than one and different from $c+1$
are
{} $R_i = \neg R_1 \wedge Q_{i-1}$
in the sequence of the lemma. Since $R_1$ is $S_1 \wedge Q_c$, this formula
{} $R_i = \neg R_1 \wedge Q_{i-1}$
is the same as
{} $R_i = \neg (S_1 \wedge Q_c) \wedge Q_{i-1}$,
which is equivalent to
{} $(\neg S_1 \wedge Q_{i-1}) \vee (\neg Q_c \vee Q_{i-1})$.
The two formulae $Q_c$ and $Q_{i-1}$ are mutually inconsistent since $i$ is not
equal to $c+1$. As a result, $Q_{i-1}$ implies $\neg Q_c$. This proves
{} $\neg Q_c \vee Q_{i-1}$
equivalent to $Q_{i-1}$. Therefore, $R_i$ in the sequence of the lemma is
equivalent to
{} $(\neg S_1 \wedge Q_{i-1}) \vee Q_{i-1}$,
which is equivalent to $Q_{i-1}$, as in the sequence of the theorem.~\qed

}

The level order
{} $R = [S_1 \wedge Q_c, Q_1, \ldots, Q_{c-1},
{}       \neg S_1 \wedge Q_c,
{}       Q_{c+1}, \ldots, Q_k]$
expresses the natural revision of a level order $[Q_1,\ldots,Q_k]$. This is the
same as naturally revising an order given as its sequence of equivalence
classes~\cite{rott-09}.

The first resulting equivalence class comprises some of the models of the
revising formula $S_1$. Namely, they are the ones in the first class that
contains some models of $S_1$. All subsequent classes comprise the remaining
models in the same relative order as before.

This correspondence of definitions was a short detour in the path from
expressing a natural order by a level order. The proof was already in place,
since both the base and the induction steps are proved.

\state{natural-levels}{theorem}{}{

\begin{theorem}
\statelabel

Every natural order is equivalent to a level order of size bounded by a
polynomial in the size of the natural order.

\end{theorem}

}{

\proof How to translate a natural order $S$ into a level order $R$ is shown by
induction.

The base is case is $S=[]$, which translates into $R=[\true]$ by
Lemma~\ref{natural-empty}.

The induction case requires a way to translate a natural order comprising at
least a formula into a level order. Let $S=[S_1,S_2,\ldots,S_m]$ be the natural
order. By the inductive assumption, $[S_2,\ldots,S_m]$ translates into a level
order $Q$. Theorem~\ref{levels-natural-levels-poly} proves $S$ equivalent to
the level order $R$:

\[
R = [S_1 \wedge Q_c,
Q_1, \ldots, Q_{c-1},
\neg S_1 \wedge Q_c,
Q_{c+1}, \ldots, Q_k]
\]

This order is larger than $R$ only by $2 \times |S_1|$.

This inductively proves that the translation is possible and produces a
sequence that is at most twice the size of the original.~\qed

}

\subsection{From lexicographic to level orders}

The reduction from lexicographic to level orders follows the same scheme as
that of natural orders: base case and induction case. Only the reduction is
shown, details are in Appendix~\ref{proofs}.

The base case proves the level order $[\true]$ equivalent to the empty
lexicographic order $[]$.

\state{lexicographic-level}{lemma}{}{}{

The following lemmas and theorems show that every lexicographic order is
equivalent to a level order. The proof scheme is the same as that of natural
orders: base case and induction case. The base case is that the level order
$[\true]$ is equivalent to the empty lexicographic order $[]$. The induction
case adds a single formula at the front of the lexicographic sequence and shows
how it changes the corresponding level order.

\begin{lemma}
\statelabel

The lexicographic order $[]$ is equivalent to the level order $[\true]$.

\end{lemma}

\proof The two definitions are:

\begin{description}

\item[level order:]

\[
i \leq j \mbox{ where } I \models S_i \mbox{ and } J \models S_j
\]

\item[lexicographic order:]

\[
S=[]
\mbox{ or }
(
	I \leq_{S_1} J
	\mbox{ and }
	(
		J \not\leq_{S_1} I
		\mbox{ or }
		I \leq_R J
	)
)
\]

\end{description}

The first definition is satisfied by every pair of models $I,J$ since both
models satisfy the first formula of the level order $[\true]$; as a result,
both $i$ and $j$ are equal to $1$, and they therefore satisfy $i \leq j$.

The second definition is satisfied because of its first part $S = []$.~\qed

}

The induction step changes the level order to maintain the equivalence when
adding a formula to the lexicographic order. Namely, prefixing a formula $S_1$
to the lexicographic order $[S_2,\ldots,S_m]$ is the same as turning the
corresponding level ordering $[Q_2,\ldots,Q_k]$ into
{} $[     S_1 \wedge Q_2, \ldots,     S_1 \wedge Q_k,
{}   \neg S_1 \wedge Q_2, \ldots,\neg S_1 \wedge Q_k]$.
This proves that every lexicographic order is equivalent to some level order.

\state{lexicographic-next}{lemma}{}{}{

The induction step changes the level order to keep it equivalent to the
lexicographic order while adding a formula at time to the latter.

\begin{lemma}
\statelabel

If $S=[S_1,S_2,\ldots,S_m]$ is a lexicographic order and $Q$ is a level order
equivalent to the lexicographic order $[S_2,\ldots,S_m]$, then $I \leq_S J$ is:

\begin{itemize}

\item true if $I \models S_1$ and $J \not\models S_1$

\item false if $I \not\models S_1$ and $J \models S_1$

\item same as $I \leq_Q J$ otherwise

\end{itemize}

\end{lemma}

\proof The definition of the lexicographic order $I \leq_S J$ is:

\[
S=[]
\mbox{ or }
(
	I \leq_{S_1} J
	\mbox{ and }
	(
		J \not\leq_{S_1} I
		\mbox{ or }
		I \leq_R J
	)
)
\]

The lemma implicitly assumes $S$ not empty. What results from removing the
case $S=[]$ is:

\[
I \leq_{S_1} J \mbox{ and } ( J \not\leq_{S_1} I \mbox{ or } I \leq_R J )
\]

The comparison $I \leq_S J$ is evaluated in the four cases where $I$ or $J$
satisfy $S_1$ or not.

\begin{description}

\item[$I \models S_1$ and $J \not\models S_1$]

The definition of $I \leq_{S_1} J$ is $I \models S_1$ or $J \not\models S_1$,
and is therefore satisfied. This is the first part of $I \leq_S J$.

Similarly, $J \leq_{S_1} I$ is $J \models S_1$ or $I \not\models S_1$. Both are
false. Therefore, $J \leq_{S_1} I$ is false. Its negation $J \not\leq_{S_1} I$
is true. The second part of $I \leq_S J$ is therefore true, being
$J \not\leq_{S_1} I \mbox{ or } I \leq_R J$

\item[$I \not\models S_1$ and $J \models S_1$]

The definition of $I \leq_{S_1} J$ is $I \models S_1$ or $J \not\models S_1$;
both conditions are false. Since $I \leq_{S_1} J$ is false, its conjunction
with $J \not\leq_{S_1} I \mbox{ or } I \leq_R J$ is also false. Since this
conjunction is equivalent to $I \leq_S J$, this comparison is false as well.

\item[$I \models S_1$ and $J \models S_1$]

The first assumption $I \models S_1$ implies $I \leq_{S_1} J$.

The second assumption $J \models S_1$ implies $J \leq_{S_1} I$.

The condition 
{} $I \leq_{S_1} J \mbox{ and } ( J \not\leq_{S_1} I \mbox{ or } I \leq_R J )$
simplifies to the equivalent condition
{} $\true \mbox{ and } ( \false \mbox{ or } I \leq_R J )$,
which is the same as $I \leq_R J$.

\item[$I \not\models S_1$ and $J \not\models S_1$]:

The first assumption $I \not\models S_1$ implies $J \leq_{S_1} I$.

The second assumption $J \not\models S_1$ implies $I \leq_{S_1} J$.

The condition 
{} $I \leq_{S_1} J \mbox{ and } ( J \not\leq_{S_1} I \mbox{ or } I \leq_R J )$
simplifies to the equivalent condition
{} $\true \mbox{ and } ( \false \mbox{ or } I \leq_R J )$,
which is the same as $I \leq_R J$.

\end{description}~\qed

The second part of the induction step is representing the order $I \leq_S J$
in the previous lemma as a level order.

}

\state{level-lexicographic-levels}{lemma}{}{}{

\begin{lemma}
\statelabel

If the lexicographic order $[S_2,\ldots,S_m]$ is equivalent to the level
order $Q=[Q_1,\ldots,Q_k]$, then the lexicographic order
$[S_1,S_2,\ldots,S_m]$ is equivalent to the following level order.

\[
R = [
S_1 \wedge Q_1,\ldots,S_1 \wedge Q_k,
\neg S_1 \wedge Q_1,\ldots,\neg S_1 \wedge Q_k
]
\]

\end{lemma}

\proof The lemma assumes that $Q$ is a level order (its formulae are mutually
inconsistent) and the disjunction of all its formulae is tautologic. The models
of every formula $Q_i$ are split among $S_1 \wedge Q_i$ and $\neg S_1 \wedge
Q_i$. Therefore, $R$ has the same properties of $Q$.

The rest of the proof shows that $S$ is equivalent to $R$.

Lemma~\ref{lexicographic-next} proves that $I \leq_S J$ is:

\begin{itemize}

\item true if $I \models S_1$ and $J \not\models S_1$

\item false if $I \not\models S_1$ and $J \models S_1$

\item same as $I \leq_Q J$ otherwise

\end{itemize}

The same is proved for $I \leq_R J$. The starting point is the definition of
level order: $I \leq_R J$ is
{} $i \leq j \mbox{ and } I \models R_i \mbox{ and } J \models R_j$.
It is evaluated in the three cases above.

\begin{description}

\item[$I \models S_1$ and $J \not\models S_1$]

Let $R_i$ and $R_j$ be the formulae of $R$ respectively satisfied by $I$ and
$J$.

Since $I$ also satisfies $S_1$ by assumption, it satisfies $S_1 \wedge R_i$.
Since $J$ falsifies $S_1$ by assumption, it satisfies $\neg S_1$ and therefore
also $\neg S_1 \wedge R_j$.

These formulae $S_1 \wedge R_i$ and $\neg S_1 \wedge R_j$ are in the positions
$i$ and $k+j$ in the sequence $R$. Since $i$ is an index a sequence of length
$k$, it holds $i \leq k$. As a result, $i < k+j$. This inequality implies $i
\leq k+j$, which defines $I \leq_S J$.

\item[$I \not\models S_1$ and $J \models S_1$]

Let $R_i$ and $R_j$ be the formulae of $R$ respectively satisfied by $I$ and
$J$. The assumptions $I \not\models S_1$ implies $I \models \neg S_1$. As a
result, $I$ satisfies $\neg S_1 \wedge R_i$. Since $J$ satisfies $J \models
S_1$ by assumption, it also satisfies $S_1 \wedge R_j$.

The formulae $\neg S_1 \wedge R_i$ and $S_1 \wedge R_j$ are in the positions
$k+i$ and $j$ in the sequence $R$. Since $j$ is an index a sequence of length
$k$, it holds $j \leq k$. As a result, $j < k+i$. This inequality is the
opposite of $k+i \leq j$, which defines $I \leq_S J$. This comparison is
therefore false, as required.

\item[otherwise]

The two remaining cases are 
{} $I \models S_1$ and $J \models S_1$
and
{} $I \not\models S_1$ and $J \not\models S_1$.

Let $R_i$ and $R_j$ be the formulae of $R$ respectively satisfied by $I$ and
$J$.

The conditions
{} $I \models S_1$ and $J \models S_1$
imply
{} $I \models S_1 \wedge R_i$ and $J \models S_1 \wedge R_j$.
These formulae are in the sequence $R$ at positions $i$ and $j$. The definition
of $I \leq_S J$ is $i \leq j$, which is also the definition of $I \leq_R J$ in
this case.

The conditions
{} $I \not\models S_1$ and $J \not\models S_1$
imply
{} $I \models \neg S_1$ and $J \models \neg S_1$,
which imply
{} $I \models \neg S_1 \wedge R_i$ and $J \models \neg S_1 \wedge R_j$.
These formulae are in the sequence $R$ at positions $k+i$ and $k+j$. The
definition of $I \leq_S J$ is $i \leq j$, which is equivalent to $k+i \leq
k+j$, which defines $I \leq_R J$ in this case.

\end{description}~\qed

This lemma shows how to keep a level order equivalent to a lexicographic order
while adding a formula to the latter.

}

\state{lexicographic-equivalent-level}{theorem}{}{

\begin{theorem}
\statelabel

Every lexicographic order is equivalent to a level order.

\end{theorem}

}{

\proof The claim is proved by induction on the length of the lexicographic
order $S$.

The base case is $S=[]$. Its equivalent level order is $R=[\true]$. It is
equivalent because both compare $I \leq J$ for all models. The first because
$S=[]$ is one of the condition of its definition. The second because both $I$
and $J$ always satisfy $\true$, the first formula of $R$.

In the induction case, the sequence $S$ has length one or more. Let
$S_1,S_2,\ldots,S_m$ be its formulae. Lemma~\ref{level-lexicographic-levels}
requires a level order $Q$ to be equivalent to $[S_2,\ldots,S_m]$; it exists by
the induction assumption. As a result, the claim of the lemma holds: $S$ is
equivalent to a level order $R$.~\qed

}

The theorem proves that every lexicographic order can be translated into a
level order, but neglects size. It does not say that the level order is
polynomial in the size of the lexicographic order. As a matter of facts, it is
not. Some lexicographic orders explode into exponentially larger level orders.
The next section proves this.

\section{Comparison}
\label{comparison}

Which representations are able to represent all doxastic states? Which do it
shortly?

\subsection{Expressivity}

In propositional logic on a finite alphabet, all considered four
representations are universal~\cite{gura-kodi-18,schw-etal-22}: each represents
all connected preorders.

The explicit representation is actually just the mathematical formalization of
a connected preorder. Every connected preorder is representable by definition.
The explicit representation is universal.

\state{explicit-level}{theorem}{}{

\begin{theorem}
\statelabel

Every connected preorder is the level ordering of a sequence of mutually
inconsistent formulae.

\end{theorem}

}{

\proof Every connected preorder corresponds to a sequence of disjoint sets $E =
[E_1,\ldots,E_m]$, where $I \leq J$ corresponds to $I \in E_i$, $J \in E_j$ and
$i \leq j$. In the specific case of propositional models, every set of models
$E_i$ is the set of models of a formula $S_i$. Therefore, a connected preorder
is also the level order $[S_1,\ldots,S_m]$.~\qed

}

The natural and lexicographic representations are proved universal indirectly:
the level representation is translated into each of them. Since the level
representation is universal, these are as well. These two translations are in
the next section.

\subsection{Compactness}

The translations from natural and lexicographic orders to level orders are in
the previous section. A translation from natural to lexicographic orders is in
a previous article~\cite{libe-23}, which however neglects size.

Since natural and lexicographic orders are defined inductively, an inductive
definition of level orders facilitates the translations.

\state{levels-inductive}{lemma}{}{

\begin{lemma}
\statelabel

It holds $I \leq_S J$ holds if and only if the following condition holds, where
$S$ is a level order, $S_1$ is its first formula and $R$ the sequence of the
following ones.

\[
S=[]
\mbox{ or } 
I \in \mod(S_1)
\mbox{ or } 
(J \not\in \mod(S_1) \mbox{ and } I \leq_R J)
\]

\end{lemma}

}{

\proof The definition of the level order $I \leq_S J$ is:

\[
\forall j . J \not\models S_j
\mbox{ or }
i \leq j \mbox{ where } I \models S_i \mbox{ and } J \models S_j
\]

This definition is proved to coincide with the condition in the statement of
the lemma.

Since the condition is ``$S=[]$ or something else{}'', it is true when $S=[]$.
This is also the case for the definition, since no formula of the sequence is
satisfied by a model $J$.

If $S$ is not empty, the condition of the statement of the lemma becomes:

\[
I \in \mod(S_1)
\mbox{ or } 
(J \not\in \mod(S_1) \mbox{ and } I \leq_R J)
\]

This is proved to coincide with the definition of the level orders in all
cases: $I$ satisfies $S_1$, it does not and $J$ does, and none of them do. The
proof is by induction: it is assumed true on sequences strictly shorter than
$S$.

\begin{description}


\item[$I \models S_1$]




The condition is true because it is ``$I \in \mod(S_1)$ or something else{}'',
and $I \in \mod(S_1)$ is true because of the assumption $I \models S_1$.

The definition is also true. The assumption $I \models S_1$ implies that $I
\models S_i$ with $i=1$. Let $S_j$ be the formula such that $J \models S_j$.
Since the sequence starts at $1$, this index $j$ is larger or equal than $1$.
Since $i$ is equal to one, $j \geq i$ follows. This is the same as $i \leq j$,
which defines $I \leq_S J$.

\item[$I \not\models S_1$ and $J \models S_1$]




The condition simplifies from
{} $I \in \mod(S_1)
{}  \mbox{ or }
{}  (J \not\in \mod(S_1) \mbox{ and } I \leq_R J)$
to $\false \mbox{ or } (\not\true \mbox{ and } I \leq_R J$, which is false.

The definition is not met either. Since $J \models S_1$, the first part of the
definition $\forall j. J \not\models S_j$ is false. Since $J \models S_1$, the
index $j$ such that $J \models S_j$ is $1$. Since $I$ does not satisfy $S_1$,
it either satisfies $S_i$ with $i > 1$ or it does not satisfy any formula of
$S$. In the first case, $j > i$ implies that $i \leq j$ is false. In the second
case, $I \models S_i$ is false for all formulae $S_i$. The second part of the
definition
{} {} $i \leq j \mbox{ where } I \models S_i \mbox{ and } J \models S_j$
is false either way.

\item[$I \not\models S_1$ and $J \not\models S_1$]




These two assumptions imply that $I \in \mod(S_1)$ and $J \in \mod(S_1)$ are
false. The condition
{} $I \in \mod(S_1) \mbox{ or } (J \not\in \mod(S_1) \mbox{ and } I \leq_R J)$
simplifies to
{} $\false \mbox{ or } (\true \mbox{ and } I \leq_R J))$,
which is equivalent to
{} $I \leq_R J$,
where $R=[S_2,\ldots,S_m]$.

The definition of the level order also simplifies.

Its first part
{} $\forall j. J \not\models S_j$
is equivalent to
{} $\forall j>1. J \not\models S_j$,
since $J$ is known not to satisfy $S_1$. This is the same as
{} $\forall j. J \not\models R_j$.

Its second part
{} $i \leq j \mbox{ where } I \models S_i \mbox{ and } J \models S_j$
may only hold with $i>1$ and $j>1$ since neither $I$ nor $J$ satisfy $S_1$. As
a result, it simplifies to
{} $i+1 \leq j+1 \mbox{ where } I \models R_i \mbox{ and } J \models R_j$
where $R=[S_2,\ldots,S_m]$. This is equivalent to
{} $i \leq j \mbox{ where } I \models R_i \mbox{ and } J \models R_j$.

The conclusion is that the definition of $I \leq_S J$ is the same as
{} $\forall j. J \not\models R_j
{}  \mbox{ or }
{}  i \leq j \mbox{ where } I \models R_i \mbox{ and } J \models R_j$,
the definition of $I \leq_R J$.

\end{description}~\qed

}

\subsubsection{From level to natural orders}

Level orders translate into natural orders in polynomial time and space: every
level order of a sequence of mutually inconsistent formulae is the natural
order of the same sequence.

The proof comprises two steps. The first is a technical result: the models that
falsify all formulae of a natural order are greater than or equal to every
other model. This is the case because the formulae of a natural order state a
belief in the truth of their models. The falsifying models are unsupported, and
therefore unbelieved.

The second step of the proof is an inductive expression of the natural order $I
\leq_S J$: it holds if $S$ is either empty, or the first formula of $S$
supports $I$, it denies $J$, or $I$ is less than or equal to $J$ according to
the rest of $S$. This expression is the same as the level order of the same
sequence.

\state{natural-no-true}{lemma}{}{}{

The next results prove that level orders translate into natural orders in
polynomial time and space.

Lemma~\ref{levels-inductive} proves that $I \leq_S J$ equates a certain
inductive condition on the level order $S$. The following theorem proves the
same for natural orders when $S$ comprises mutually inconsistent formulae. For
these sequences the identity translates level orders into equivalent natural
orders. This suffices since level orders can be restricted to mutually
inconsistent formulae.

A preliminary lemma on natural orders is necessary.

\begin{lemma}
\statelabel

If a model $J$ falsifies all formulae of the natural order $S$, then $I \leq_S
J$ holds for every model $I$.

\end{lemma}

\proof The definition of $I \leq_S J$ is:

\begin{eqnarray*}
&& S = [] \mbox{ or } \\
&&
(I \in \mod(S_1) \mbox{ and } \forall K \in \mod(S_1) . I \leq_R K)
\mbox{ or } \\
&&
	I \leq_R J
	\mbox{ and }
		(
		J \not\in \mod(S_1)
		\mbox{ or }
		\exists K \in \mod(S_1) . J \not\leq_R K
		)
)
\end{eqnarray*}

In the base case $S=[]$ this definition is met. As a result, $I \leq_S J$ holds
by definition.

The induction case is proved as follows. Since $J$ does not satisfy any formula
of $S=[S_1,S_2,\ldots,S_m]$, it does not satisfy $S_1$ and does not satisfy any
formula of $R=[S_2,\ldots,S_m]$. The latter implies $I \leq_R J$ by the
induction assumption. The definition of $I \leq_S J$ simplifies as follows
when replacing $J \not\in \mod(S_1)$ and $I \leq_R J$ by $\true$.

\begin{eqnarray*}
\lefteqn{S = [] \mbox{ or }}
\\
&&
(I \in \mod(S_1) \mbox{ and } \forall K \in \mod(S_1) . I \leq_R K)
\\
&&
\mbox{ or }
(
	I \leq_R J
	\mbox{ and }
		(
		J \not\in \mod(S_1)
		\mbox{ or }
		\exists K \in \mod(S_1) . J \not\leq_R K
		)
)
\\
&\equiv&
\false
\mbox{ or }
(I \in \mod(S_1) \mbox{ and } \forall K \in \mod(S_1) . I \leq_R K)
\mbox{ or }
\\
&&
(
	\true
	\mbox{ and }
		(
		\true
		\mbox{ or }
		\exists K \in \mod(S_1) . J \not\leq_R K
		)
) \\
&\equiv&
(I \in \mod(S_1) \mbox{ and } \forall K \in \mod(S_1) . I \leq_R K)
\mbox{ or }
(
	\true
	\mbox{ and }
	\true
) \\
&\equiv&
(I \in \mod(S_1) \mbox{ and } \forall K \in \mod(S_1) . I \leq_R K)
\mbox{ or }
\true
\\
\true
\end{eqnarray*}~\qed

This lemma allows expressing a natural order in the same way of a level order
when its formulae are mutually inconsistent.

}

\state{natural-inconsistent}{lemma}{}{}{

\begin{lemma}
\statelabel

If $S$ is a sequence of mutually inconsistent formulae, the natural order $I
\leq_S J$ holds if and only if the following condition holds, where $S_1$ is
the first formula of $S$ and $R$ the sequence of the following formulae of $S$.

\[
S=[]
\mbox{ or } 
I \in \mod(S_1)
\mbox{ or } 
	(J \not\in \mod(S_1) \mbox{ and } I \leq_R J)
\]

\end{lemma}

\proof Since the formulae of $S$ do not share models, if a model $K$ satisfies
$S_1$ it falsifies all other formulae $S_2,\ldots,S_m$. The latter implies $I
\leq_R K$ by Lemma~\ref{natural-no-true} since $R=[S_2,\ldots,S_m]$. In other
words, $I \leq_R K$ holds for all formulae that satisfy $S_1$. In formulae,
{} $\forall K \in \mod(S_1) . I \leq_R K$.

For the same reason,
{} $\forall K \in \mod(S_1) . J \leq_R K$
holds as well.
This is the contrary of
{} $\exists K \in \mod(S_1) . J \not\leq_R K$,
which is therefore false.

Replacing
{} $\forall K \in \mod(S_1) . I \leq_R K$
with $\true$ and
{} $\exists K \in \mod(S_1) . J \not\leq_R K$
with $\false$ in the definition of the natural order $I \leq_S J$ yields:

\begin{eqnarray*}
\lefteqn{S = [] \mbox{ or }}
\\
&&
(I \in \mod(S_1) \mbox{ and } \forall K \in \mod(S_1) . I \leq_R K)
\\
&&
\mbox{ or }
(
	I \leq_R J
	\mbox{ and }
		(
		J \not\in \mod(S_1)
		\mbox{ or }
		\exists K \in \mod(S_1) . J \not\leq_R K
		)
)
\\
&\equiv&
S = []
\mbox{ or }
(I \in \mod(S_1) \mbox{ and } \true)
\mbox{ or }
(
	I \leq_R J
	\mbox{ and }
		(
		J \not\in \mod(S_1)
		\mbox{ or }
		\false
		)
)
\\
&\equiv&
S = []
\mbox{ or }
I \in \mod(S_1)
\mbox{ or }
(
	I \leq_R J
	\mbox{ and }
	J \not\in \mod(S_1)
)
\end{eqnarray*}

The final condition is the claim of the lemma.~\qed

}

\state{levels-natural-identity}{corollary}{}{}{

Since both level orders and natural orders are equivalent to the same
condition when their formulae are mutually inconsistent, they are equivalent.

\begin{corollary}
\statelabel

Every level order of a sequence of mutually inconsistent formulae is the
natural order of the same sequence.

\end{corollary}

The translation is the identity. It takes polynomial time and space. This
concludes the proof.

}

\begin{theorem}
\label{levels-lexicographic-polyspace}

Level orders translate into natural orders in polynomial time and space.

\end{theorem}

\subsubsection{From level to lexicographic orders}

That level orders translate to lexicographic orders is a consequence of the
translation from level to natural orders shown above and the translation from
natural to lexicographic orders proved in a previous article~\cite{libe-23}.
Yet, the latter translation is not polynomial in time. What shown next is one
that is.

The translation is the identity: every level order of a sequence of mutually
inconsistent formulae is the lexicographic order of the same sequence. This is
proved by showing that the lexicographic order $I \leq_S J$ holds if and only
if either $S$ is empty, its first formula makes $I$ true, or it makes $J$ false
or the rest of the order compares it greater than or equal to $I$. This is the
same as the expression of level orders proved by Lemma~\ref{levels-inductive}.

\state{lexicographic-no-true}{lemma}{}{}{

The following results prove that level orders translate into lexicographic
order in polynomial time and space. This is proved by showing that every level
order of a sequence of mutually inconsistent formulae is the lexicographic
order of the same sequence.

The first step is a preliminary lemma.

\begin{lemma}
\statelabel

If a model $J$ falsifies all formulae of the sequence $S$, then the
lexicographic ordering $I \leq_S J$ holds for every model $I$.

\end{lemma}

\proof The definition of the lexicographic order $I \leq_S J$ is:

\[
S = []
\mbox{ or }
(
	I \leq_{S_1} J
	\mbox{ and }
	(
		J \not\leq_{S_1} I
		\mbox{ or }
		I \leq_R J
	)
)
\]

This definition is the disjunction of $S=[]$ with another condition. The base
case of induction $S=[]$ therefore meets the definition.

In the induction case, $S=[]$ is false. The definition reduces to:

\[
I \leq_{S_1} J
\mbox{ and }
(
	J \not\leq_{S_1} I
	\mbox{ or }
	I \leq_R J
)
\]


Since $J$ does not satisfy any formula of $S$, it does not satisfy its first
formula $S_1$. In turn, $J \not\models S_1$ imply $I \models S_1$ or $J
\not\models S_1$, the definition of $I \leq_{S_1} J$. Since $J$ does not
satisfy any formula of $S$, it does not satisfy any formula of its subsequence
$R$. By the induction assumption, $I \leq_R J$ holds. The definition of the
lexicographic order further simplifies to:

\[
\true
\mbox{ and }
(
	J \not\leq_{S_1} I
	\mbox{ or }
	\true
)
\equiv
\true
\mbox{ and }
\true
\equiv
\true
\]~\qed

}

\state{inconsistent-lexicographic}{lemma}{}{}{

This lemma allows proving the equivalent condition.

\begin{lemma}
\statelabel

If $S$ is a sequence of mutually inconsistent formulae, the lexicographic order
$I \leq_S J$ holds if and only if the following condition holds, where $S_1$ is
the first formula of $S$ and $R$ the sequence of the following formulae of $S$.

\[
S=[]
\mbox{ or } 
I \in \mod(S_1)
\mbox{ or } 
	(J \not\in \mod(S_1) \mbox{ and } I \leq_R J)
\]

\end{lemma}

\proof The condition in the statement of the lemma is proved to coincide with
the definition of the lexicographic order $I \leq_S J$:

\[
S = []
\mbox{ or }
(
	I \leq_{S_1} J
	\mbox{ and }
	(
		J \not\leq_{S_1} I
		\mbox{ or }
		I \leq_R J
	)
)
\]

Both the definition and the condition are disjunctions comprising $S=[]$.
Therefore, they are both true in the base case of induction $S=[]$.

In the induction case, $S$ is not empty. The definition and the condition
respectively became:

\begin{eqnarray*}
&& 
I \leq_{S_1} J
\mbox{ and }
(
	J \not\leq_{S_1} I
	\mbox{ or }
	I \leq_R J
)
\\
&&
I \in \mod(S_1)
\mbox{ or } 
	(J \not\in \mod(S_1) \mbox{ and } I \leq_R J)
\end{eqnarray*}

The induction assumption implies that the two occurrences of $I \leq_R J$
coincide since $R$ is strictly shorter than $S$.

Two cases are considered: either $I$ satisfies $S_1$ or not.

\begin{description}

\item[$I \models S_1$]

The condition in the statement of the lemma is true since it is a disjunction
comprising $I \in \mod(S_1)$.

The assumption $I \models S_1$ implies $I \leq_{S_1} J$ by definition. It also
simplifies the definition of $J \leq_{S_1} I$ from $J \models S_1$ or $I
\not\models S_1$ into just $J \models S_1$.

The definition of $I \leq_S J$ therefore simplifies:

\begin{eqnarray*}
\lefteqn{
I \leq_{S_1} J
\mbox{ and }
(
	J \not\leq_{S_1} I
	\mbox{ or }
	I \leq_R J
)
}
\\
&\equiv&
\true
\mbox{ and }
(
	J \not\models S_1
	\mbox{ or }
	I \leq_R J
)
\\
&\equiv&
J \not\models S_1
\mbox{ or }
I \leq_R J
\end{eqnarray*}

This is true if $J \not\models S_1$, and is now proved true in the other case,
$J \models S_1$.

Since the formulae $S_i$ do not share models and $J$ is a model of $S_1$, it is
not a model of any other formula $S_2,\ldots,S_m$. These formulae make $R$: no
formula of $R$ is satisfied by $J$. Lemma~\ref{lexicographic-no-true} proves $I
\leq_R J$.

Since the definition of $I \leq_S J$ is equivalent to 
{} $J \not\leq_{S_1} I \mbox{ or } I \leq_R J$,
it is met.

\item[$I \not\models S_1$]

This assumption implies $J \leq_{S_1} I$, which makes $J \not\leq_{S_1} I$
false. Since $I \leq_{S_1} J$ is defined as $I \models S_1$ or $J \not\models
S_1$, it becomes the same as $J \not\models S_1$.

The definition of $I \leq_S J$ simplifies:

\begin{eqnarray*}
\lefteqn{
I \leq_{S_1} J
\mbox{ and }
(
	J \not\leq_{S_1} I
	\mbox{ or }
	I \leq_R J
)
} \\
&\equiv&
J \not\models S_1
\mbox{ and }
(
	\false
	\mbox{ or }
	I \leq_R J
)
\\
&\equiv&
J \not\models S_1
\mbox{ and }
I \leq_R J
\end{eqnarray*}

The condition in the statement of the lemma also simplifies thanks to the
current assumption $I \not\models S_1$:

\begin{eqnarray*}
\lefteqn{
I \in \mod(S_1)
\mbox{ or } 
	(J \not\in \mod(S_1) \mbox{ and } I \leq_R J)
} \\
&\equiv&
\false
\mbox{ or } 
	(J \not\in \mod(S_1) \mbox{ and } I \leq_R J)
\\
&\equiv&
J \not\models S_1 \mbox{ and } I \leq_R J
\end{eqnarray*}

This is the same as the definition of $I \leq_S J$.

\end{description}~\qed

}

\state{level-lexicographic-identity}{corollary}{}{}{

The translation is a corollary.

\begin{corollary}
\statelabel

Every level order of a sequence of mutually inconsistent formulae is the
natural order of the same sequence.

\end{corollary}

Since the translation is the identity, it takes linear space and time. This
concludes the proof that level orders translate into lexicographic orders in
polynomial time and space.

}

\begin{theorem}

Level orders translate into lexicographic orders in polynomial time and space.

\end{theorem}

\subsubsection{From natural to lexicographic orders}

This translation follows from two previous results:
Theorem~\ref{natural-levels} shows a polynomial translation from natural to
level orders; Corollary~\ref{levels-lexicographic-polyspace} show the same from
level to lexicographic orders.

\begin{theorem}

Natural orders translate into lexicographic orders in polynomial space.

\end{theorem}

\subsection{From lexicographic to level and natural orders}

All three representations are universal: they express all connected preorders.
Therefore, they translate to each other. Whether they do in polynomial time or
space is another story. What proved next is that not only polynomiality is
unattainable in time, but also in space: some lexicographic orders are only
equivalent to exponentially long natural orders.

The troublesome lexicographic orders are not even complicated:
$[x_1,\ldots,x_n]$ is an example. The equivalence classes of this lexicographic
order contain one model each. Therefore, they are exponentially many. The
equivalence classes of level and natural orders are bounded by their number of
formulae. Exponentially many classes equal exponentially many formulae.

The proof comprises two parts: the classes of lexicographic orders may be
exponentially many; they never are for level and natural orders. Level orders
are first.

\state{level-number-classes}{lemma}{}{

\begin{lemma}
\statelabel

The level order of a sequence of $m$ formulae has at most $m+1$ equivalence
classes.

\end{lemma}

}{

\proof As shown in Section~\ref{definitions}, a level order is equivalent to
another where every model is satisfied by exactly one formula by the addition
of a single formula. This is an increase from $m$ to $m+1$ formulae.

In this other order, the condition $\forall j . J \not\models S_j$ is always
false. As a result, $I \leq_S J$ holds if and only if $i \leq j$, where $I
\models S_i$ and $J \models S_j$. The reverse comparison $J \leq_S I$ holds if
and only if $j \leq i$. As a result, $I$ and $J$ are compared the same if and
only if $i = j$, where $I \models S_i$ and $J \models S_j$. Two models are
equivalent if and only if they satisfy the same formula of $S$. This is an
isomorphism between the equivalence classes and the formulae of the
sequence.~\qed

}

Combining this result with the translation of
Theorem~\ref{levels-natural-levels-poly} proves the same statement for natural
orders.

\state{natural-number-classes}{lemma}{}{

\begin{lemma}
\statelabel

The natural order of a sequence of $m$ formulae has at most $m+1$ equivalence
classes.

\end{lemma}

}{

\proof Theorem~\ref{levels-natural-levels-poly} translates a natural order into
an equivalent level order by iterating over the formulae of the sequence, each
time turning a sequence $Q$ into a sequence $R$:

\[
R = [S_1 \wedge Q_c,
Q_1, \ldots, Q_{c-1},
\neg S_1 \wedge Q_c,
Q_{c+1}, \ldots, Q_k]
\]

The formulae of $R$ are the same of $Q$ except for the absence of $Q_c$ and the
presence of $S_1 \wedge Q_c$ and $\neg S_1 \wedge Q_c$. The number of formulae
therefore increases by one at each step. Since the process starts from $Q=[]$
and iterates over the $m$ formulae of the natural order, it produces a level
order of the same length.

By Lemma~\ref{level-number-classes}, this level order has at most $m+1$
equivalence classes. Since the natural order is equivalent, it has the same
equivalent classes.~\qed

}

The second part of the proof is that the lexicographic order
{} $[x_1,\ldots,x_n]$
comprises $2^n$ equivalence classes. A preliminary lemma is necessary.

\state{lexicographic-all}{lemma}{}{

\begin{lemma}
\statelabel

The lexicographic comparisons $I \leq_S J$ and $J \leq_S I$ hold at the same
time only if $I \leq_{S_k} J$ and $J \leq_{S_k} I$ both hold for every formula
$S_k$ of $S$.

\end{lemma}

}{

\proof The claim is that $I \leq_S J$ and $J \leq_S I$ imply $I \leq_{S_k} J$
and $J \leq_{S_k} I$ for every formula $S_k$ of $S$.

The definition of $I \leq_S J$ is:

\[
S=[]
\mbox{ or }
(
	I \leq_{S_1} J
	\mbox{ and }
	(
		J \not\leq_{R_1} I
		\mbox{ or }
		I \leq_R J
	)
)
\]

The claim is proved by induction. If $S$ is the empty sequence, the conclusion
is vacuously true since $S$ contains no formula $S_k$.

The inductive assumption is that the claim holds for every sequence $R$ shorter
than $S$. The premise of the lemma is that both $I \leq_S J$ and $J \leq_S I$
hold. The claim can be split in three parts: $I \leq_{S_1} J$, $J \leq_{S_1}
I$, and the same for every $S_k$ with $k > 1$.

In the inductive case $S$ is not empty. The definition of $I \leq_S J$ becomes:

\[
I \leq_{S_1} J
\mbox{ and }
(
	J \not\leq_{R_1} I
	\mbox{ or }
	I \leq_R J
)
\]

Since $I \leq_S J$ holds by the premise of the lemma, its first conjunct $I
\leq_{S_1} J$ holds as well. For the same reason, $J \leq_S I$ implies $J
\leq_{S_1} I$. This proves the first two parts of the claim.

Since $I \leq_S J$ holds by the premise of the lemma, its second conjunct
{} $J \not\leq_{R_1} I \mbox{ or } I \leq_R J$
holds as well. Its first disjunct is contradicted by $J \leq_{S_1} I$, proved
above. What is left is $I \leq_R J$. The same argument proves that $J \leq I$
implies $J \leq_R I$.

By the induction assumption, $I \leq_R J$ and $J \leq_R I$ imply $I \leq_{S_k}
J$ and $J \leq_{S_k} I$ for every formula $S_k$ of $R$. These are the formulae
$S_k$ of $S$ with $k > 1$.~\qed

}

The number of equivalence classes of 
{} $[x_1,\ldots,x_n]$
can now be proved.

\state{lexicographic-variables-classes}{lemma}{}{

\begin{lemma}
\statelabel

The lexicographic order $S=[x_1,\ldots,x_n]$ has $2^m$ equivalence classes.

\end{lemma}

}{

\proof Two different models differs on a variable $x_k$ at least:
{} $I \models x_k$ and $J \not\models x_k$.

If $I$ and $J$ were equivalent according to $S$, then
Lemma~\ref{lexicographic-all} would apply. It implies
{} $I \leq_{x_k} J$ and $J \leq_{x_k} I$
for every $k$. The second conclusion $J \leq_{x_k} I$ is the same as $J \models
x_k$ or $I \not\models x_k$, both of which are false. Therefore, $I$ and $J$
are not equivalent according to $S$.

The conclusion is that two different models are not equivalent. Every model is
in its own class of equivalence. Since the models are $2^m$, the classes of
equivalence are the same number.~\qed

}

This result negates translations from lexicographic orders to level and natural
orders of polynomial size.

\state{lexicographic-exponential-level}{theorem}{}{

\begin{theorem}
\statelabel

The lexicographic order $[x_1,\ldots,x_n]$ is only equivalent to level and
natural orders comprising at most $2^n-1$ formulae.

\end{theorem}

}{

\proof The lexicographic order $[x_1,\ldots,x_n]$ is assumed equivalent to a
level or natural order $S$ by contradiction. By
Lemma~\ref{lexicographic-variables-classes}, the lexicographic order
$[x_1,\ldots,x_n]$ has $2^n$ equivalence classes. Since the level or natural
order $S$ is equivalent to it, it has the same equivalence classes. By
Lemma~\ref{level-number-classes} or Lemma~\ref{natural-number-classes}, $S$
comprises at least $2^m-1$ formulae.~\qed

}

\draft

conditionally skipped:
introduction to the problem of efficiency in revising
the various representations of the state

\enddraft

\skipif{}{y}
[subsection]

\subsection{operations}

The comparison between the representations of the epistemic states is done
regarding their expressivity (how many connected preorder they represent) and
their compactness (how small their representations of connected preorder are).
An open direction of study is their efficiency: how fast orders are operated
upon:

\begin{itemize}

\item how much does it cost to check $I \leq_S J$?

\item how much does it cost to revise or contract an order? how large the
resulting order is?

\end{itemize}

An obvious result of the second kind is that naturally revising a natural order
is linear in time: revising $[S_1,\ldots,S_m]$ by a new formula $P$ produces
$[P,S_1,\ldots,S_m]$. The same operation may not be that easy on lexicographic
orders.

These open questions are left for future work.

The present one continues on the study of compactness. How different
representation compare on this measure is identified. The winners are the
lexicographic orders. All other orders translate to them in polynomial time and
therefore size, while some lexicographic order do not in the other way around.
Yet, lexicographic orders may be large. The same connected preorder may be
represented by several lexicographic orders, some small and some large.

Revising a lexicographic order $[S_1,\ldots,S_m]$ by a new formula $P$ produces
$[P,S_1,\ldots,S_m]$. This sequence may be equivalent to smaller ones. This is
not just a matter of size. It is also a matter of assimilating information.
Remembering how every piece of information has come be known is unrealistic. At
some point, information is restructured, summarized, generalized. Learned
rather than remembered.

A specific case of learning is redundancy elimination. Some piece of
information may come several times. Some other may come when they are already
known. Some others may be later widened. The general question is whether a
lexicographic order may be removed formulae without changing its meaning.

The same question applies to all representation of epistemic states. It studied
for lexicographic orders in the next section.

[vedere altro da dire in rankingrevision.txt]

\skipfi

\section{Related work}
\label{related}

Most work in the iterated belief revision literature are purely semantical, but
computational aspects are not neglected. An early example is the work by
Ryan~\cite{ryan-91}, who wrote: ``Belief states are represented as deductively
closed theories. This means that they are (in general) impossible to write down
fully, or to store on a computer{}''; he employed a partial order between a
finite number of formulae to represent the doxastic state.
%
%
Williams~\cite{will-92} and Dixon~\cite{dixo-93} represented doxastic states by
ordered partitions of formulae.
%
%
%
Williams~\cite{will-95} later introduced partial entrenchment rankings,
functions from a set of formulae to integers.
%
%
Dixon and Wobcke~\cite{dixo-wobc-93} observed: ``it is not possible to
represent all entrenchments directly: some entrenchments allow infinitely many
degrees of strength of beliefs. Moreover, it is impossible for the user of a
system to enter all entrenchment relations: a more compact representation must
be found{}''; their solution is to allow for a partial specification, an
ordered partition of formulae.
%
%

Computational issues are kept into account rarely~%
\cite{benf-etal-99,benf-etal-00,jin-thie-05,zhua-etal-07,rott-09},
%
%
%
%
%
%
but recently attracted interest~%
\cite{gura-kodi-18,souz-etal-19,saue-hald-19,%
schw-etal-20,arav-20,saue-beie-22,schw-etal-22}.
%
%
%
%
%
%
%
%
Most solutions employ structures equivalent or analogous to
level orders~%
\cite{will-92,dixo-93,will-95,dixo-wobc-93,%
jin-thie-05,meye-etal-02,rott-09,schw-etal-20},
lexicographic orders~%
\cite{ryan-91,benf-etal-00,zhan-04},
or histories of revisions~%
\cite{benf-etal-99,schw-etal-22};
the history of revisions may also be necessary for semantical, rather than
computational, reasons~%
\cite{koni-pere-00,roor-etal-03,hunt-delg-07}.
Some other solutions change or extend these three solutions, and some others
steer away from them.
Souza, Moreira and Vieira~\cite{souz-etal-19} employ priority
graphs~\cite{liu-11}, strict partial orders over a set of formulae.
Aravanis~\cite{arav-20} follow Areces and Becher~\cite{arec-bech-01} in their
semantics based on a fixed ordering on the models.
%


Schwind, Konieczny and Pino P\'erez~\cite{schw-etal-22} introduced a concept of
doxastic state equivalence: ``two epistemic states are strongly equivalent
according to a revision operator if they cannot be distinguished from each
other by any such successive revision steps, which means that these epistemic
states have the same behavior for that revision operator{}''. This abstract
definition generalizes Definition~\ref{equivalence} to representations that are
not connected preorders among propositional models.
%
%

\section{Conclusions}
\label{conclusions}

\draft
{\bf SUMMARY OF THE ARTICLE}
\enddraft

How large a doxastic state is? It depends on how it is stored. Four ways are
considered: explicit, by a list of formulae expressing equivalence classes and
by a history of revisions, either lexicographic or natural.

To compare different representation, they are defined inductively. The
definitions for the history of revisions are shown equivalent to the
definitions based on equivalence classes in the literature, while showing at
the same time how they can be converted into the representation by equivalence
classes.

The comparison is completed by investigating the other reductions, both their
existence and their compactness, their ability to store doxastic states in
little space. All four representations are universal: each represents all
possible connected preorders on propositional models. They radically differ on
compactness. The explicit representation is the more wasteful: it is always
exponential in the number of state variables, unlike the others. The
representation by equivalence classes and by natural revisions are more
compact, and equally so. The most compact of the four representations is that
by lexicographic revisions. The other three representations can always be
converted into it with a polynomial increase in size, while the converse
reduction may produce exponentially large results.

\draft
{\bf FUTURE DIRECTIONS}
\enddraft

Investigation can proceed in many directions. Compactness does not only matter
when comparing different representations. It also matters within the same. The
same doxastic state has multiple lexicographic representations, for example.
Some are short, some are long. The question is whether one can be made shorter.
This problem is similar to Boolean formulae minimization~%
\cite{mccl-56,rude-sang-87,coud-94,theo-etal-96,coud-sasa-02,uman-etal-06}.
A related question is how to shrink a connected preorder below a certain size
while minimizing the loss of information. A subcase of interest is revision
redundancy, whether a revision can be removed from a history without changing
the resulting doxastic state.

The level, lexicographic and natural representations of the doxastic states are
the most common in iterated belief revisions, but others are possible. An
example is a single formula over an alphabet $Y \cup Z$ that is true on a model
$I$ if and only if $I[X/Y]$ is less than or equal than $I[X/Z]$. Other
representations of the doxastic state have been proposed~%
\cite{arav-etal-18,andr-etal-02,gura-kodi-18,souz-etal-19},
%
%
%
%
%
such as prioritized bases~%
\cite{brew-89,nebe-91,benf-etal-93},
weighted knowledge bases~%
\cite{bouz-etal-14,etta-etal-23}
and conditionals~%
\cite{kuts-19,andr-etal-02,saue-etal-22}.
The preference reasoning field offers many alternatives~%
\cite{doms-etal-11}.
An order among models may not suffice~%
\cite{boot-chan-17,boot-chan-20}.
Similar representation issues also arise in belief merging~%
\cite{diaz-pino-23}.
%


\appendix
\section{Proofs}
\label{proofs}


\restate{natural-empty}
\restate{first-consistent}
\restate{natural-next}
\restate{level-natural-level}
\restate{levels-natural-levels-poly}
\restate{natural-levels}

\restate{lexicographic-level}
\restate{lexicographic-next}
\restate{level-lexicographic-levels}
\restate{lexicographic-equivalent-level}


\restate{explicit-level}

%

\restate{levels-inductive}

\restate{natural-no-true}
\restate{natural-inconsistent}
\restate{levels-natural-identity}

\restate{lexicographic-no-true}
\restate{inconsistent-lexicographic}
\restate{level-lexicographic-identity}


\restate{level-number-classes}
\restate{natural-number-classes}
\restate{lexicographic-all}
\restate{lexicographic-variables-classes}
\restate{lexicographic-exponential-level}

\bibliographystyle{alpha}
\newcommand{\etalchar}[1]{$^{#1}$}

\end{document}